\documentclass[final]{cvpr}

\usepackage{times}
\usepackage{epsfig}
\usepackage{graphicx}
\usepackage{amsmath}
\usepackage{amssymb}
\usepackage{cite}
\usepackage{threeparttable}
\usepackage{multirow}
\usepackage{booktabs}
\usepackage{soul}
\usepackage{amsmath,amsfonts,bm}

\def\eqref#1{equation~\ref{#1}}
\def\1{\bm{1}}

\DeclareMathAlphabet{\mathsfit}{\encodingdefault}{\sfdefault}{m}{sl}
\SetMathAlphabet{\mathsfit}{bold}{\encodingdefault}{\sfdefault}{bx}{n}

\newcommand{\R}{\mathbb{R}}

\usepackage{subfigure}
\usepackage{floatrow}

\usepackage[pagebackref=true,breaklinks=true,colorlinks,bookmarks=true]{hyperref}

\def\cvprPaperID{3259} % *** Enter the CVPR Paper ID here
\def\confYear{CVPR 2021}
\newcommand{\name}[0]{ResizeMix}

\begin{document}

%%%%%%%%% TITLE
\title{\name: Mixing Data with Preserved Object Information and True Labels}

\author{Jie Qin$^{1,2}$\footnotemark[1] , Jiemin Fang$^{3,4}$\footnotemark[1] , Qian Zhang$^{5}$, Wenyu Liu$^{4}$, Xingang Wang$^{2}$\footnotemark[2] , Xinggang Wang$^{4}$\vspace{5pt}\\
$^1$School of Artificial Intelligence, University of Chinese Academy of Sciences\\
$^2$Institute of Automation, Chinese Academy of Sciences\\
$^3$Institute of Artificial Intelligence, Huazhong University of Science and Technology\\
$^4$School of EIC, Huazhong University of Science and Technology $\; ^5$Horizon Robotics\\
{\tt\small \{qinjie2019, xingang.wang\}@ia.ac.cn} \; {\tt\small qian01.zhang@horizon.ai}\\
{\tt\small \{jaminfong, liuwy, xgwang\}@hust.edu.cn}
}

\maketitle

%%%%%%%%% ABSTRACT
\begin{abstract}
Data augmentation is a powerful technique to increase the diversity of data, which can effectively improve the generalization ability of neural networks in image recognition tasks. Recent data mixing based augmentation strategies have achieved great success. Especially, CutMix uses a simple but effective method to improve the classifiers by randomly cropping a patch from one image and pasting it on another image. To further promote the performance of CutMix, a series of works explore to use the saliency information of the image to guide the mixing. We systematically study the importance of the saliency information for mixing data, and find that the saliency information is not so necessary for promoting the augmentation performance. Furthermore, we find that the cutting based data mixing methods carry two problems of \textbf{label misallocation} and \textbf{object information missing}, which cannot be resolved simultaneously. We propose a more effective but very easily implemented method, namely ResizeMix. We mix the data by directly resizing the source image to a small patch and paste it on another image. The obtained patch preserves more substantial object information compared with conventional cut-based methods. \name\ shows evident advantages over CutMix and the saliency-guided methods on both image classification and object detection tasks without additional computation cost, which even outperforms most costly search-based automatic augmentation methods.
\end{abstract}
\renewcommand{\thefootnote}{\fnsymbol{footnote}}
\footnotetext[1]{Equal contribution. The work was performed during the internship of J. Qin and J. Fang at Horizon Robotics.}
\footnotetext[2]{Corresponding author.}
\renewcommand{\thefootnote}{\arabic{footnote}}

%%%%%%%%% BODY TEXT
\section{Introduction}
\begin{figure}[t]
   \centering
   \subfigure[CutMix with Object Information Missing and Label Misallocation]{\label{fig: cutmix}
      \includegraphics[width=1.0\linewidth]{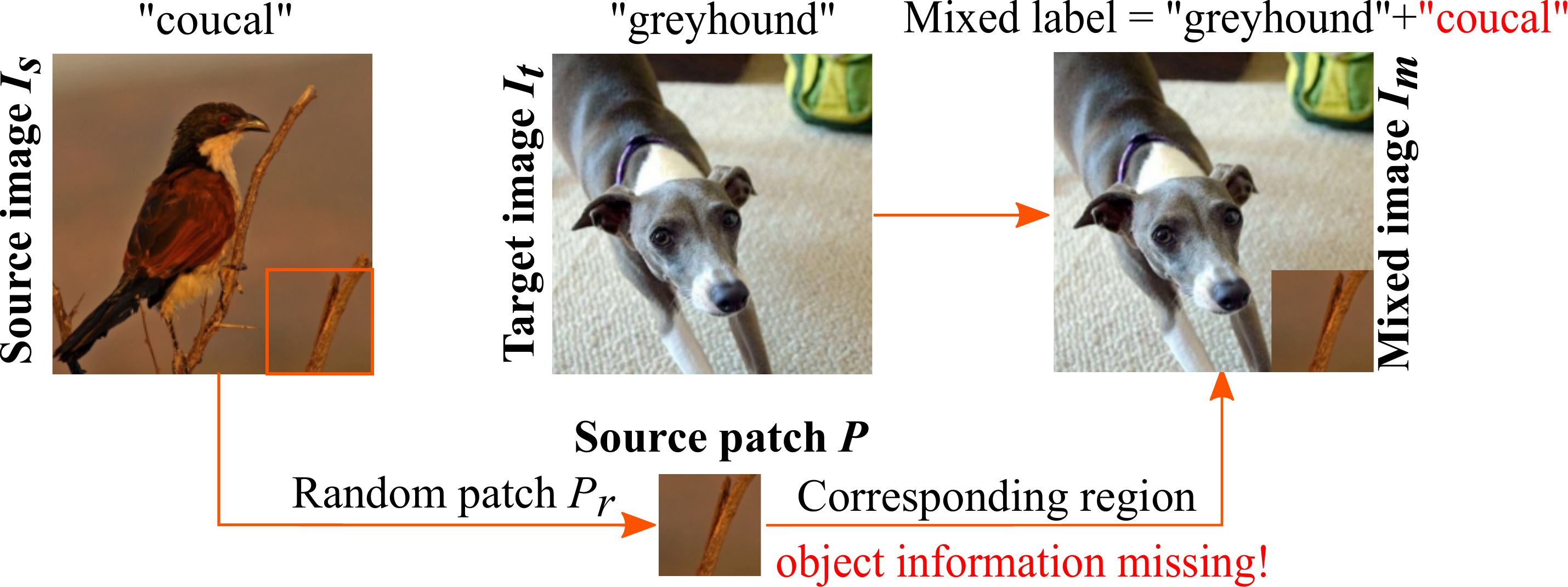}}
      \subfigure[Possible Choices of Data Mixing]{
      \label{fig: checking}
      \includegraphics[width=1.0\linewidth]{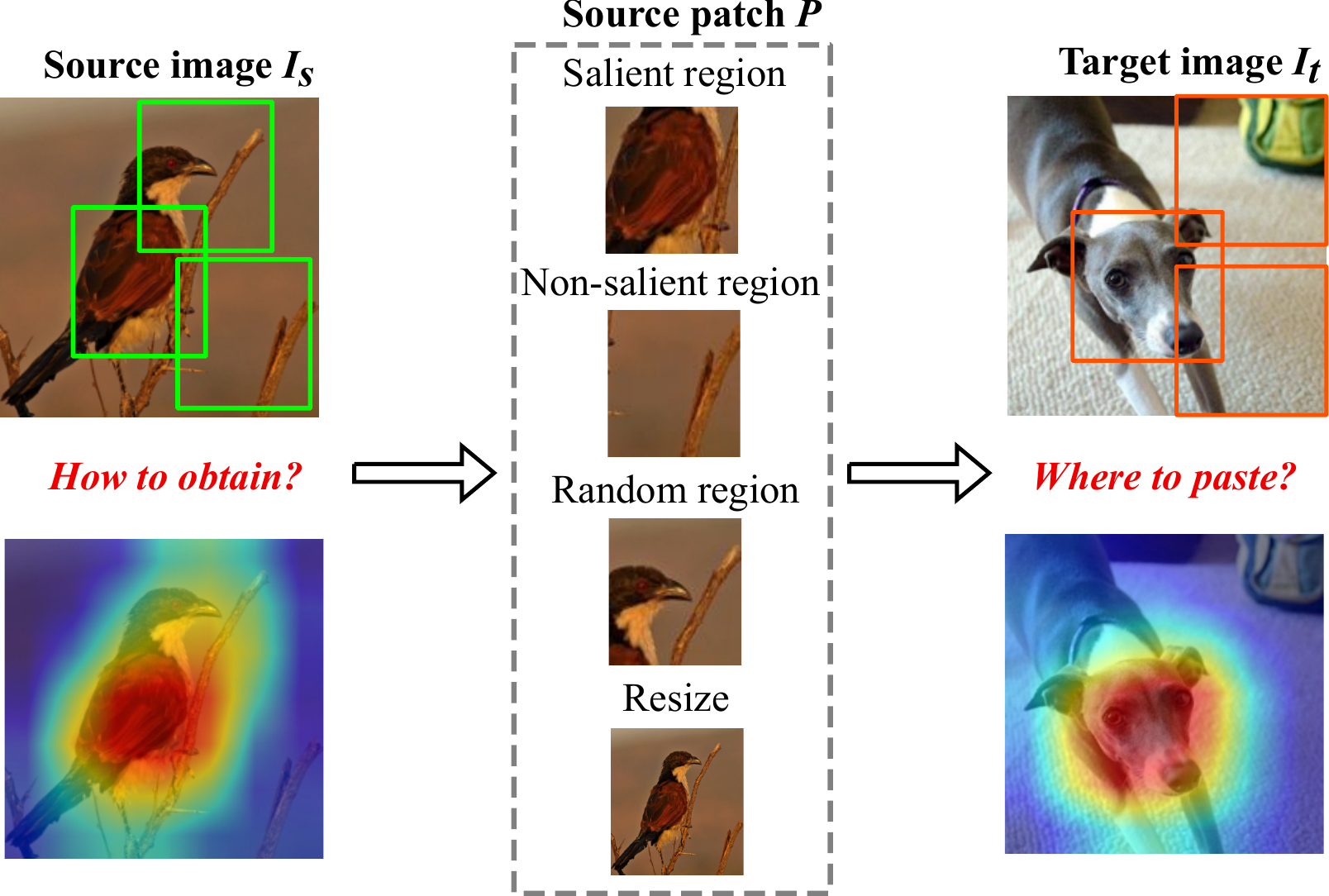}}
      \caption{(a) illustrates that in CutMix~\cite{cutmix}, there exist two issues of object information missing and label misallocation. (b) represents different cropping manners from the source image and different pasting manners to the target image. We systematically check the two problems of ``how to obtain the patch'' and ``where to paste the patch''.\vspace{-15pt}}
   \end{figure} 

Deep convolutional neural networks (CNN) have achieved great success in a wide range of computer vision applications, \eg, image classification~\cite{he2016resnet,sandler2018mobilenetv2}, object detection~\cite{ren2015faster,lin2017focal}, and semantic segmentation~\cite{ronneberger2015u,DBLP:journals/pami/ChenPKMY18} \etc. Recent optimization techniques have further promoted the CNN performance to a new level, including data augmentation~\cite{autoaugmentcubuk,cutmix}, optimizer design~\cite{DBLP:journals/corr/KingmaB14,DBLP:journals/corr/Alistarh0TV16}, learning rate schedule~\cite{DBLP:conf/iclr/LoshchilovH17}, and hyper-parameter optimizing~\cite{bergstra2012random,domhan2015speeding} \etc. Notably, the strategy of data augmentation plays a critically important role in broadening the distribution of data, which facilitates the generalization ability of the trained CNN, and effectively promotes the final performance. Advanced data augmentation methods have been widely explored for training stronger neural networks.

A series of data augmentation methods~\cite{mixup,summers2019improved,cutmix} aim at mixing data to increase the data diversity. Meanwhile, the mixed data forces the network to pay attention to multiple objects and locations in the input image, which strengthens the feature extraction ability of the networks. Especially, as shown in Fig.~\ref{fig: cutmix}, CutMix~\cite{cutmix} achieves very promising results on image classification by randomly cutting a \emph{source patch} from a \emph{source image} and pasting it on the \emph{target image} at the same location. The ground truth labels are accordingly mixed proportionally to the patch area, which leads to a multi-label training style. However, the mixing strategy with a random manner may mislead the training, as the cropped patch usually does not conform to the label of the whole image. Subsequent researches~\cite{kim2020puzzlemix,harris2020fmix,uddin2020saliencymix} make efforts to mix the data more precisely, most of which take full advantages of the saliency information and use the location of the salient regions in the image to guide the mixing. The saliency-guided method facilitates the consistency of the mixed data and the allocated ground truth labels. This alleviates the misleading caused by the random mixing strategy during training, and further promotes the neural network performance. 

However, the procedure of locating the salient region of the image always requires a complicated module and introduces additional computation cost during training, \eg, PuzzleMix~\cite{kim2020puzzlemix} proposes to optimize the mixing mask and the saliency discounted transportation, and SaliencyMix~\cite{uddin2020saliencymix} uses a saliency detection module to select the saliency source patch for mixing. In this paper, we systematically check the importance of the image saliency information for data mixing during network training. As shown in Fig.~\ref{fig: checking}, the checking is performed mainly from two aspects, \ie whether the saliency information is necessary for determining \emph{(\romannumeral1) where to paste the source patch} and \emph{(\romannumeral2) how to obtain the source patch}. 

For evaluating the two questions, we employ a Grad-CAM~\cite{selvaraju2017gradcam} module to locate the salient region in the image, and perform a series of studies about the saliency information for mixing. As a consequence, for (\romannumeral1), we find that the saliency-guided location surpasses that in CutMix, which keeps the location consistent in the two mixing images; while randomly determining the pasting location further surpasses the saliency-guided location. This indicates that the saliency information indeed facilitates the pasting location determining, but is defeated by the random location in terms of the data diversity. For (\romannumeral2), the cropped patch from the salient region only achieves similar performance with the randomly cropped patch. How to obtain a better image patch for mixing still remains an unsolved question. As shown in Fig.~\ref{fig: cutmix}, we deduce the cutting manner for obtaining the image patch is easy to cause \emph{label misallocation} due to the semantic inconsistency between the cropped patch and the whole source image, and \emph{object information missing} which is verified in our experiment. Based on the above clues, we propose a novel and effective data mixing method, namely \name, which directly resizes the image and pastes the resized patch on another image. \name\ eliminates the label misallocation issue and preserves substantial information for mixing. The proposed \name\ consistently outperforms CutMix~\cite{cutmix} and latter saliency-guided methods~\cite{kim2020puzzlemix,harris2020fmix,uddin2020saliencymix} on both CIFAR and ImageNet classification tasks. When transferred to the MS-COCO object detection task, the model trained on ImageNet with \name\ shows evident advantages over CutMix.

We summarize our contributions as follows.
\begin{enumerate}
\vspace{-5pt}
    \item Considering saliency information is widely used in recent mixing-based augmentation methods, we systematically check the importance of the saliency information, and find that saliency information is not so necessary for mixing data.
\vspace{-5pt}
    \item We verify that cropping the patch for mixing is easy to cause label misallocation and object information missing, and propose a new mixing method \name, which resolves the two issues by directly resizing the image for mixing.
\vspace{-5pt}
    \item The proposed \name\ shows evident advantages over CutMix and the saliency-guided methods on both image classification and object detection tasks without any additional computation cost, which even outperforms most costly search-based automatic augmentation methods.
\end{enumerate}

%------------------------------------------------------------------------
\section{Related Work}

% Data augmentation is a powerful technique for machine learning to improve the generalization performance of their models by increasing the amount and quality of image samples. Traditionally, geometric transformations and color spaced transformations have been used in data augmentation, such as random cropping, random flipping, and color-shifting were used to improve the accuracy of classification tasks~\cite{classification, deep}. These methods required extra time and experience to design params, to alleviate these shortcomings, many convenient methods were proposed to improve the performance of models, such as the cutting augmentations~\cite{erazing2020, cutout2017, improved}, the mixing  augmentations~\cite{mixup, between, cutmix} and the automated augmentations~\cite{autoaugmentcubuk, fastaalim, pba, ratner2017learning}. Herein, we first introduce some cutting and mixing augmentations in Sec.~\ref{section: cut and mix}. Then we retrospect the recent development of saliency-guided mixing augmentations in Sec.~\ref{section: saliencymix}. Finally, we review the progress of automated augmentations systematically in Sec.~\ref{section: autoaug}.

\paragraph{Cutting- and Mixing- based Data Augmentation}
\label{section: cut and mix}
The goal of cutting augmentations is to make a network pay attention to the entire data like the dropout regularization~\cite{srivasdropout2014, faster2015towards,choe2019attention, 2018dropblock, 2017hide}. Random erasing~\cite{erazing2020} selects a patch of an image and masks it out. The width and height of the patch need to be designed manually. Beyond this, Cutout~\cite{cutout2017} proposes to mask a region with a fixed-size square. Another type of augmentation methods are based on mixing data. Mixup~\cite{mixup} attempts to produce an element-wise convex combination of two images. Augmix~\cite{hendrycks2019augmix} mixes up the images augmented by operations sampled from the spaces like AutoAugment~\cite{autoaugmentcubuk} defined ones. Rather than mixing the element-wise convex, RICAP~\cite{takahashi2018ricap} randomly gets four patches from different images and combines them to a new sample. CutMix~\cite{cutmix} randomly crops a patch from one image and pastes it into the corresponding position of another image, which significantly improves the test accuracy and exceeds most augmentation methods on various datasets.  

\vspace{-6pt}
%-------------------------------------------------------------------------
\paragraph{Saliency Guided Data Augmentation}
\label{section: saliency mix}
Recently, mixing-based augmentation methods are widely used to augment images because they do not require extra searching or training cost while bringing significant performance improvement of networks. For example, CutMix~\cite{cutmix} significantly improves the test accuracy and exceeds the most automatic augmentation methods~\cite{autoaugmentcubuk,fastaalim,cubuk2020randaugment}. However, the cropping and pasting method may cause label misallocation when the cropped patch is from the background of the image. Some studies further improve the performance of CutMix by reserving patches with more saliency information when cropping and pasting the patch between two images. PuzzleMix~\cite{kim2020puzzlemix}, which proposes to optimize the position of the mixing mask and the saliency discounted transportation. SuperMix~\cite{dabouei2020supermix} uses the knowledge of a teacher to mix images on their salient regions. Somewhat differently, FMix~\cite{harris2020fmix} sets a threshold for the low-frequency parts in the image to get the saliency masks for mixing images.  SaliencyMix~\cite{uddin2020saliencymix} uses a saliency detection module to select the saliency source patch for mixing. However, they all need extra cost to find the saliency regions. Compared to these methods, we propose a convenient and effective approach that can preserve the object information of images.

\vspace{-6pt}
\paragraph{Automated Data Augmentation}
\label{section: autoaug}
Parallel with the success of neural architecture search~\cite{zoph2018learning,liu2018darts,cai2018proxylessnas,fang2020densely,fang2020fast,fang2020fna++}, automated augmentation methods start to develop rapidly. AutoAugment~\cite{autoaugmentcubuk} attempts to search for better combinations of augmentation operations and their magnitudes. Due to its expensive search cost when implemented with reinforcement learning, PBA~\cite{pba} with population evolution strategy and FastAA~\cite{fastaalim} with matching density are proposed to speed up training without reducing the performance. The augmentation combinations can be treated as a hyper-parameter optimization formulation. OHL-AA~\cite{lin2019online} tries to optimize the probability distribution of augmentations, while Faster AA~\cite{fasteraahataya2019} and DADA~\cite{li2020dada} use the differentiable optimization directly to search the combinations and magnitudes of augmentations, which can save lots of searching cost. Integrated with adversarial training~\cite{antoniou2017data,gurumurthy2017adversarial, peng2018jointlyadersarial}, AdvAA~\cite{zhang2019adversarialaa} makes networks learn more hard data samples, in which the domain of dataset becomes more widespread. Different from the search or optimization strategies, RandAugment~\cite{cubuk2020randaugment} reaches identical performance only set up two parameters with the same augmentation spaces. Overall, most of the automated augmentation methods need extra search or training cost to obtain better performance, while our proposed \name\ can promote the network performance without any additional cost.

\section{Checking the Importance of Saliency Information for Mixing Data}
\label{sec: checking}
In this section, we systematically check whether the saliency information is necessary for mixing data. First, we introduce the preliminaries for our checking process in Sec.~\ref{section: preliminaries}. Then we check the importance of saliency information from two perspectives, \ie where to paste the source patch in Sec.~\ref{section: where_paste} and how to obtain the source patch in Sec.~\ref{section: how_obtain}.

\subsection{Preliminaries}
\label{section: preliminaries}
We use $I_s \in \R^{W \times H}$ and $I_t \in \R^{W \times H}$ to denote the source and target image respectively. We denote the source patch obtained from the source image as $P \in \R^{W_P \times H_P}$, while the patch cropped from the salient region as $P_{s}$, from the non-salient region as $P_{ns}$, and from a random region as $P_{r}$. 
% For simplicity, we assume $W = H$ and $W_P = H_P$.

We employ a Grad-CAM~\cite{selvaraju2017gradcam} module to obtain the salient and non-salient pixels in the image by calculating the heatmap. The Grad-CAM module is connected to the end of the backbone network. 
Specifically, $C_{s}$ represents a set of salient pixel coordinates where the activation value of the heatmap is greater than a certain upper threshold $t_{u}$; on the contrary, $C_{ns}$ represents a non-salient coordinate set where the activation value is under a lower threshold $t_{l}$. They are defined as 
\begin{equation}
    \label{equation: regions}
    \begin{aligned}
    C_s &= \{(x, y) | A(x, y) \ge t_{u}\}\text{,}\\
    C_{ns} &= \{(x, y) | A(x, y) \le t_{l}\}\text{,}
    \end{aligned}
\end{equation}
where $(x, y)$ denotes the coordinate of a pixel in the image, and $A(x, y)$ denotes the activation value at the position of $(x,y)$. 
% $R_{r}$ represents the whole pixel position of the image.
% We use $W$ and $H$ to denote the width and height of the whole image, $W_P$ and $H_P$ to denote the width and height of the source patch. For simplicity, we assume $W = H$ and $W_P = H_P$.

\begin{table}[t!]
   \centering
   \caption{Checking results on CIFAR-100 with WideResNet-28-10 about different manners of obtaining and the locations to paste the source patch. The column of ``Type'' means how the source patch is obtained, cutting or resizing from the source image. The ``Region'' column indicates the region in the source image to generate the source patch. The ``Pasting Region'' column indicates the location in the target image to paste the source patch.}
   \label{tab: saliency_check}\small
   \begin{threeparttable}
   \resizebox{\linewidth}{!}{
        \begin{tabular}{cccccc}
        \toprule
        \multirow{2}{*}{\textbf{Row}} & \multicolumn{2}{c}{\textbf{Source Patch}} & \multirow{2}{*}{\textbf{\begin{tabular}{c}
           Pasting\\Region
        \end{tabular}}} & \multirow{2}{*}{\textbf{\begin{tabular}{c}
        Top-1 \\ Acc(\%)
        \end{tabular}}} \\ 
        & \textbf{Type} & \textbf{Region} & \\
        \midrule
        (1) Baseline & - & - & - & 81.20 \\
        \midrule
        (2) CutMix~\cite{cutmix} & Cut & Random & Corresponding & 83.40 \\
        (3) & Cut & Random & Non-salient & 83.93 \\
        (4) & Cut & Random & Salient & 83.97 \\
        (5) & Cut & Random & Random & \textbf{84.14} \\
        \midrule
        (6) & Cut & Non-salient & Random & 83.93 \\
        (7) & Cut & Salient & Random & 84.07 \\
        (8) & Cut & Random & Random & 84.14 \\
        (9) \name & Resize & Whole & Random & \textbf{84.31} \\
        \bottomrule
        \end{tabular}
        }
     \end{threeparttable}
   \end{table}   

\begin{figure}[t]
   \centering
      \includegraphics[width=0.8\linewidth]{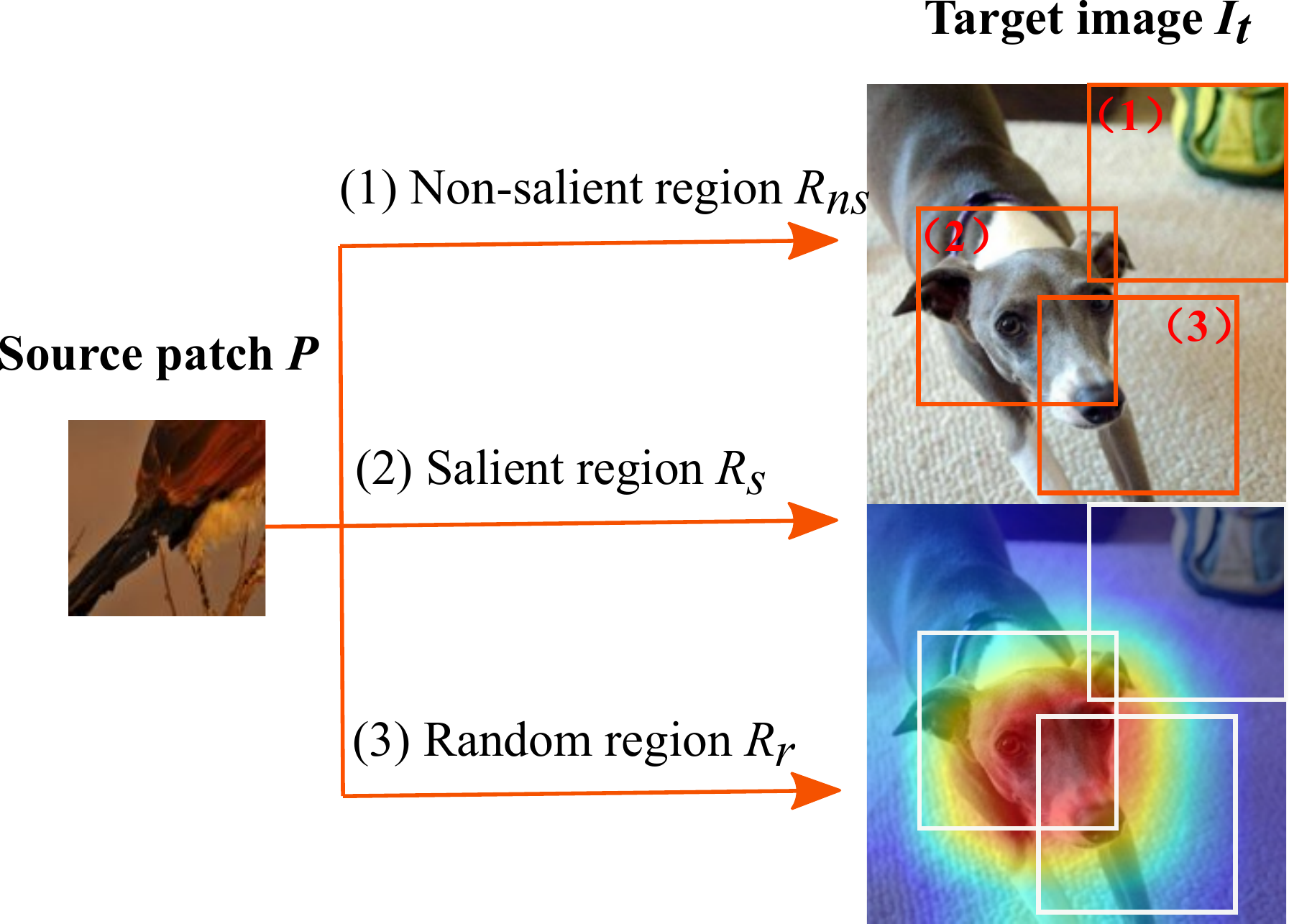}
      \vspace{-5pt}
      \caption{Three different regions to paste the source patch, including the non-salient, salient, and random region.\vspace{-10pt}}
      \label{fig: pastepatch}
   \end{figure}

We use $R(x_l, x_r, y_b, y_t)$ to denote a region of the image, and $x_l, x_r, y_b, y_t$ represent the left, right, bottom and top boundaries of the region. A salient region is denoted as $R_{s}$, whose geometric center is a salient pixel sampled from $C_s$. A non-salient region is denoted as $R_{ns}$, whose center is a non-salient pixel sampled from $C_{ns}$. And we use $R_{r}$ to denote a random region whose center is randomly sampled from the whole image. The relationship of the geometric center $(x_c,y_c)$ and the region boundaries is as follows,
\begin{equation}
    x_c=\frac{x_l+x_r}{2}\text{,\quad}
    y_c=\frac{y_d+y_u}{2}\text{.}
\end{equation}

We define the operation of pasting the source patch $P$ to the region $R(x_l, x_r, y_b, y_t)$ in the target image $I_t$ as $Paste(P, I_t, R)$,
\begin{equation}
   Paste(P, I_t, R): I_t[R] = I_t[x_l:x_r, y_b:y_t] = P \text{.}
\end{equation}
% We first randomly select a pair of coordinates $(x_s, y_s)$ from the salient region $R_s$ or the other two regions. Then we center on this pair of coordinates to locate a rectangle that owns the width $W_P$ and the height $H_P$. Finally, we make the pixel value of $\mathbf{M}$ equal to one which is in the area of the rectangle and other pixels equal to zero. This procedure is defined as follows
% \begin{equation}
%     \label{equation: mask}
%   \mathbf{M_{s}(x, y)} = \begin{cases}
%       1\text{,} \ \ \ & x\in(x_s - W_P/2\text{,}\  x_s + W_P/2) \\ 
%         \ \ \ \ \ & y\in(y_s - H_P/2\text{,}\  y_s + H_P/2) \\
%       0\text{,} \ \ \ & x\not\in(x_s - W_P/2\text{,}\  x_s + W_P/2) \\ 
%       \ \ \ \ \ & y\not\in(y_s - H_P/2\text{,}\  y_s + H_P/2)
%       \end{cases}
% \end{equation}
CutMix~\cite{cutmix} randomly crops a patch from the source image and pastes it to the target image. It can be formulated as:
\begin{equation}
   \begin{aligned}
      P &= I_s[R_r]\text{,} \\
      I_m &= Paste(P, I_t, R_r)\text{,} \\
      l_m &=\lambda l_{s}+(1-\lambda) l_{t}\text{,}
      \end{aligned}
\end{equation}
where $R_r=(x_l, x_r, y_b, y_t)$ denotes a random region where to crop the source patch and to paste on the target image. The region $R_r$ is the same in the source and target image. $l_s$, $l_t$ and $l_m$ denote the ground truth labels of the source, target and mixed image respectively. $\lambda$ is computed as the ratio of the source patch and the target image, which is formulated as,
\begin{equation}
   \label{equation: label_mix}
      \lambda = \frac{W_P * H_P}{W * H}\text{.}
\end{equation}

Since the random cropping manner proposed in CutMix may obtain the patch from the background of the image, which leads to label misallocation. To alleviate the shortcoming, some works~\cite{kim2020puzzlemix,uddin2020saliencymix, harris2020fmix} make use of the image saliency information to guide the mixing process. Considering the saliency information obtaining is usually complicated and costly, we systematically check the importance of the saliency information for mixing-based data augmentation from the following two aspects.

\subsection{Checking Saliency Information for ``Where to Paste the Source Patch''}
\label{section: where_paste}
To study the location for pasting the source patch, we crop the source patch from a random region of the source image as $P_r$, and paste it on various regions in the target image. Fig.~\ref{fig: pastepatch} shows three different kinds of locations: (1) non-salient region $R_{ns}$, (2) salient region $R_s$, and (3) random region $R_r$. 

The results of mixing data with three different patch pasting locations are shown in Row (3)-(5) of Tab.~\ref{tab: saliency_check}. We observe that the results of pasting the source patch to the non-salient region in Row (3) and the salient region in Row (4) both surpass the result of CutMix~\cite{cutmix} in Row (2). The source patch is paste to the corresponding location of the target image in CutMix. The salient or non-salient region are both more diverse than the unique corresponding location, which leads to more various mixed images. It is notable that the settings of the salient and non-salient region show similar results. This indicates the network can always extract a part of information from the target image. Therefore, the saliency information for where to paste the source patch is not so necessary. Row (5) is the result of pasting the random patch to a random region in the target image, which further surpasses both the salient and non-salient region guided ones. This indicates the random region has more diversity for mixing the images, which contains both the salient and non-salient regions.

\subsection{Checking Saliency Information for ``How to Obtain the Source Patch''}
\label{section: how_obtain}
\begin{figure}[t]
\centering
    \includegraphics[width=0.8\linewidth]{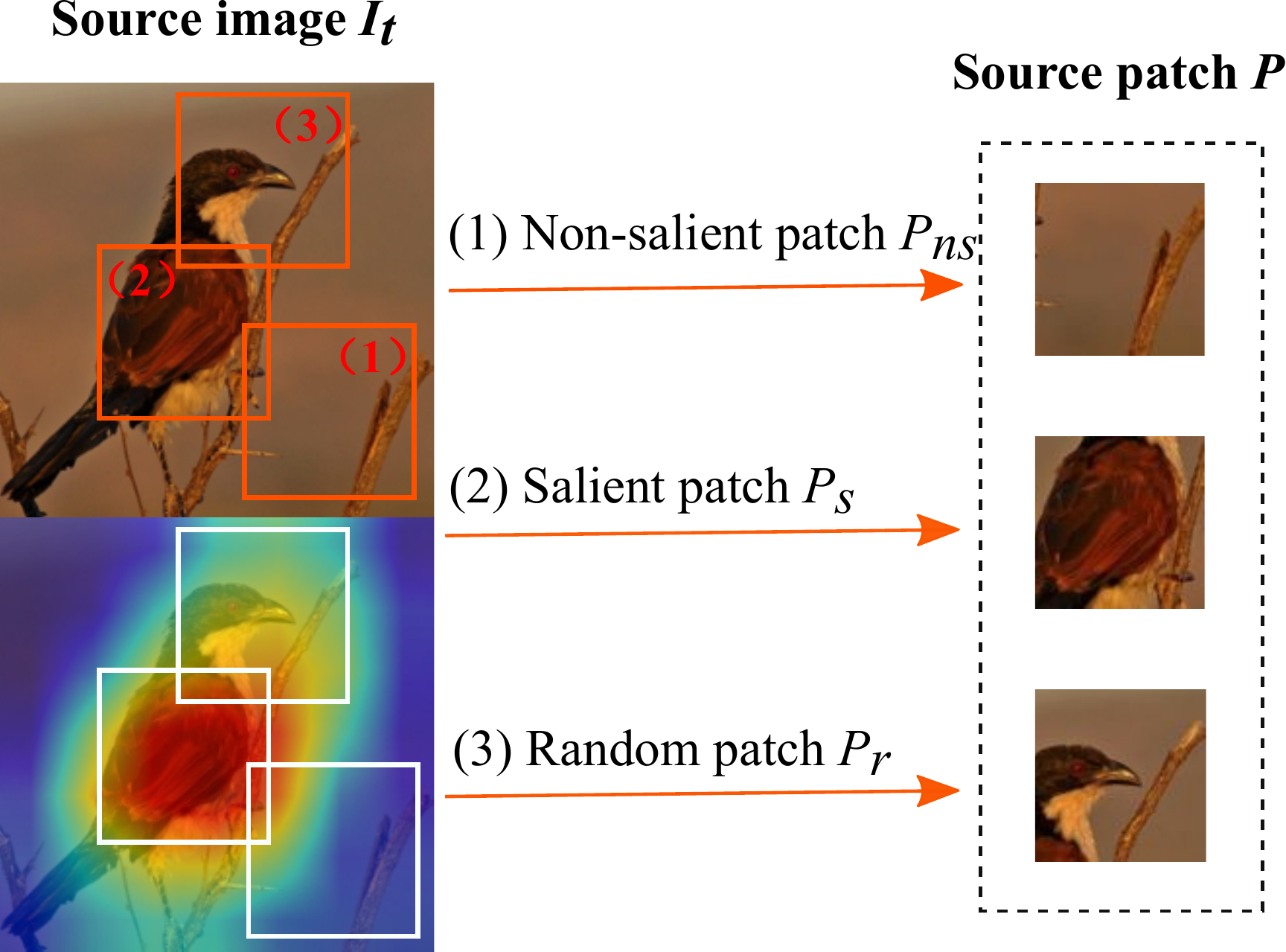}
\vspace{-5pt}
\caption{Three different manners of obtaining the source patch.
\vspace{-10pt}}
\label{fig: obtainpatch}
 \end{figure}

In this section, we check whether saliency information is necessary for obtaining the source patch from the source image. As shown in Fig.~\ref{fig: obtainpatch}, we crop patches from three different regions of the source image, \ie the salient, non-salient and random region. The source patch is paste to a random region $R_r$ of the target image.

Row (6)-(8) in Tab.~\ref{tab: saliency_check} show the results of three different types of the source patch obtaining. We find that the result of the salient patch in Row (7) surpasses the non-salient patch in Row (6). This is because the salient patch contains more information of the object corresponding to the allocated label than the non-salient patch. And if the non-salient patch contains too little information of the labeled object, this patch will lead to the problem of label misallocation. The salient patch is less possible to cause the misallocation. However, the result in Row (8) with the source patch cropped from a random region further outperforms the salient patch setting. This illustrates the random cropping manner can cover more regions with the labeled object preserved, while the salient patch only focuses on a smaller region; thus the random patch leads to more data diversity and achieves a better result. However, the random cropping strategy still carries the issue of label misallocation, how to better obtain the source patch remains an unsolved problem.

%-------------------------------------------------------------------------
\begin{figure}[t!]
   \centering
   \begin{center}
       \includegraphics[width=0.9\linewidth]{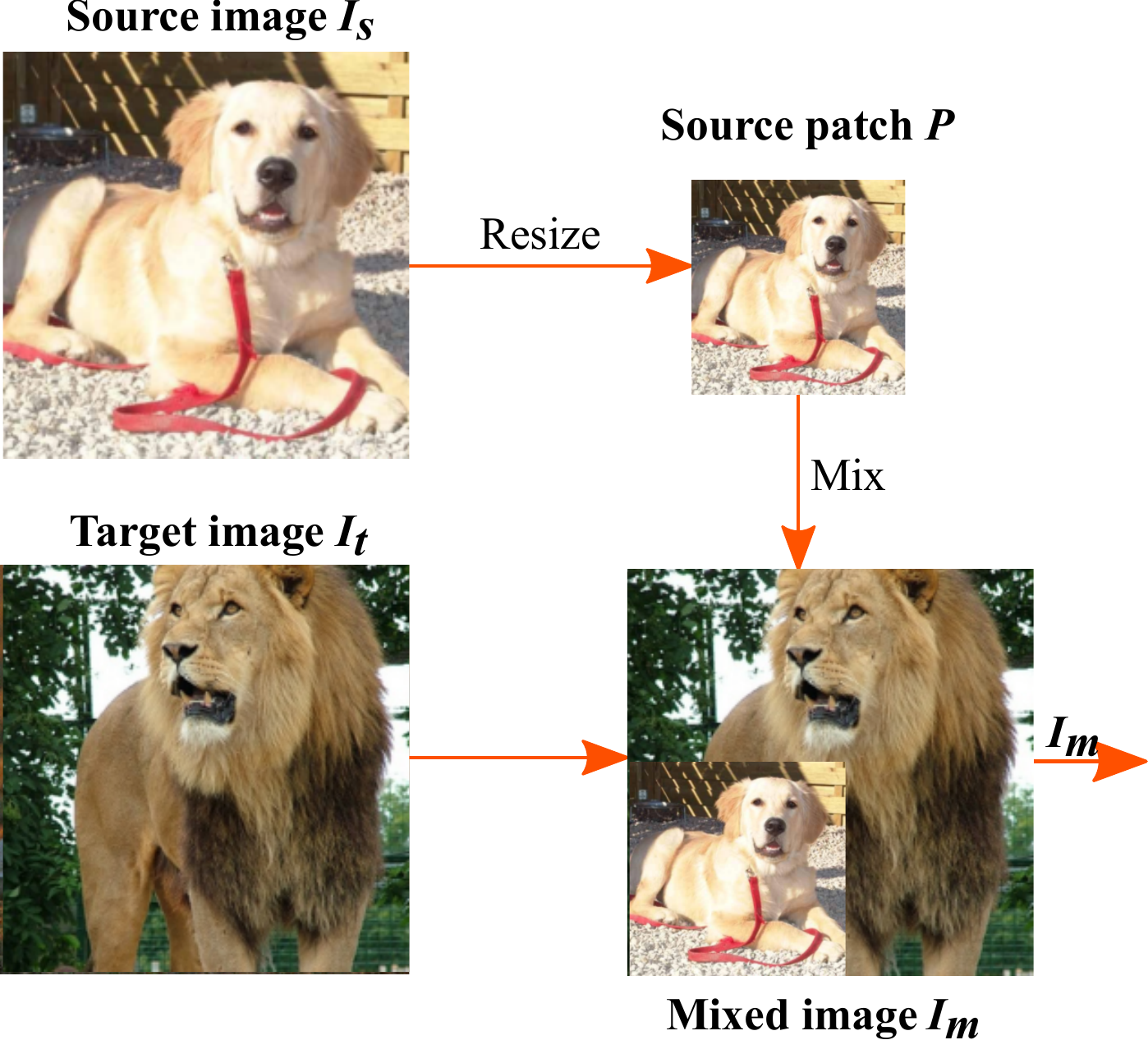}
   \end{center}
   \vspace{-8pt}
   \caption{Process of \name. The source image is resized to a smaller patch and the patch is pasted to the target image, which generates the mixed image.\vspace{-10pt}}
   \label{fig: Resizemix}
\end{figure}

\section{ResizeMix}
\label{section: resizemix}
Based on the checking results in Sec.~\ref{sec: checking}, we observe that pasting the source patch in a random region of the target image leads to the best performance. For the cutting-based strategy of obtaining the source patch, when the cutting location covers more parts of a image, the label misallocation is aggravated as some patches contain no labeled object; when the cutting location focuses on the salient region to avoid label misallocation, the diversity of the mixed image decreases and some information of the source image is lost. The two issues of label misallocation and object information missing cannot be solved simultaneously under the cutting-based strategy.
% It is obvious that cropping a patch from a source image can lead to object information missing and along with the label misallocation problem shown in Sec.~\ref{section: where_paste} and Sec.~\ref{section: how_obtain}. 

To tackle the above two problems, we propose a new data mixing method \name. As shown in Fig.~\ref{fig: Resizemix}, we directly resize the whole source image to a smaller scale as the source patch instead of cropping a patch from a local region. Then the source patch is pasted to a random region of the target image. \name\ can avoid the problem of label misallocation while the complete object information of the source image is preserved. 

Specifically, we first resize the source image $I_s$ to a smaller sized patch $P$ by a scale rate of $\tau$, which is defined as 
\begin{equation}
   P = T(I_s)\text{,}
\end{equation}
where $T()$ denotes the resizing operation and the scale rate $\tau$ is sampled from the uniform distribution $\tau \sim U(\alpha, \beta)$, where $\alpha$ and $\beta$ denote the lower and upper bound of the range respectively. Then we paste the resized patch $P$ into a random region $R_r$ in the target image. This mixing operation introduces no additional computation cost, as the scale rate and the pasting region are both obtained randomly. The image mixing is formulated as
\begin{equation}
   I_m = Paste(P, I_t, R_r)\text{.}
\end{equation}

We mix the source image label $l_s$ and the target image label $l_t$ according to the image mixing ratio $\lambda$,
\begin{equation}
   l_m = \lambda l_s + (1 - \lambda) l_t\text{,}
\end{equation}
where $\lambda$ is defined by the size ratio of the patch and the target image, \ie $\lambda = \frac{W_P * H_P}{W*H}$. $W$, $H$ and $W_P$, $H_P$ denote the width and height of the target image and the source patch respectively. As $P$ is resized from the source image with the scale rate of $\tau$, the relationship of $W$ and $W_P$ is $W_P=\tau* W$; the same as $H$ and $H_P$. Therefore, $\lambda$ and $\tau$ satisfy:
\begin{equation}
   \lambda = \tau^2\text{.}
\end{equation}

\section{Experiments}

In this section, we first study the effect of \name\ on image classification in Sec.~\ref{section: image_classification}. Then, we evaluate the generalization ability of the model pre-trained on ImageNet with \name\ by applying it on object detection in Sec.~\ref{section: detection}. Finally, we conduct some ablation studies and analysis in Sec.~\ref{section: ablation_study}.

\subsection{Evaluation on Image Classification}
\label{section: image_classification}
We evaluate the performance of \name\ on image classification dataset including CIFAR-10~\cite{krizhevsky2009learning}, CIFAR-100~\cite{krizhevsky2009learning} and ImageNet~\cite{russa2015imagenet}.
% For a fair comparison, we use the traditional augmentation setting such as resizing, cropping, and flipping as done in~\cite{han2017deep, huang2017densely}.

\subsubsection{Experiments on CIFAR-10}
The CIFAR-10 dataset contains 60,000 color images of 32$\times$32 size with 10 classes. There are 50,000 images for training and 10,000 images for validation. We implement \name\ on two neural netowrks, \ie WideResNet-28-10~\cite{zagoruyko2016wide} and Shake-Shake (26 2x96d)~\cite{gastaldi2017shake}. We train the WideResNet-28-10 network for 200 epochs with a batch size of 256 using the stochastic gradient descent (SGD) optimizer. We use the Nesterov momentum~\cite{dozat2016incorporating} of 0.9, and the weight decay of $5\times 10^{-4}$. The initial learning rate is 0.1 and decays with the cosine annealing schedule~\cite{DBLP:conf/iclr/LoshchilovH17}. When training the Shake-Shake (26 2x96d) network, we set the total epochs as 1,800 and the batch size as 256 using the SGD optimizer. The initial learning rate is 0.01 and the weight decay is $1 \times 10^{-3}$. We set the parameters of $\alpha$, $\beta$ for limiting the resizing scale ratios defined in Sec.~\ref{section: resizemix} as 0.1 and 0.8, which are used for determining the range of the patch resizing scale. 

\begin{table}[t!]
\centering
\caption{Top-1 test accuracy rate (\%) on CIFAR-10 classification with WideResNet-28-10~\cite{zagoruyko2016wide} (WRS28-10) and Shake-Shake (26 2x96d)~\cite{gastaldi2017shake} (SS-2$\times$96d). ``\name+'' denotes \name\ equipped with RandAugment~\cite{cubuk2020randaugment}. ``Cost'' represents the additional computation cost introduced by searching or adjusting augmentation strategies, and $^\dagger$ denotes the cost estimated according to the description in the original paper. ``GHs'': GPU Hours.} 
\label{tab: cifar10_results}
\begin{threeparttable}\small
      \begin{tabular}{lccc}
      \toprule
      \textbf{Method} & \textbf{Cost (GHs)} & \textbf{WRS28-10} & \textbf{SS-2$\times$96d} \\ 
      \midrule
      Baseline & 0 & 96.13 & 97.14 \\
      AA~\cite{autoaugmentcubuk} & 5000 & 97.32 & 98.00 \\
      Fast AA~\cite{fastaalim} & 3.5 & 97.30 & 98.00 \\
      PBA~\cite{pba} & 5 & 97.42 & 97.97 \\
      OHL-AA~\cite{lin2019online} & 83.4$^\dagger$ & 97.39 & - \\
      RA~\cite{cubuk2020randaugment} & 0 & 97.30 & 98.00 \\
   %  Adv AA~\cite{zhang2019adversarialaa} & 98.10 & 98.15 \\
      Faster AA~\cite{fasteraahataya2019} & 0.23 & 97.40 & 98.00 \\
      DADA~\cite{li2020dada} & 0.1 & 97.30 & 98.00 \\
      \midrule
      Cutout~\cite{cutout2017} & 0 & 96.90 & 97.14 \\
      CutMix~\cite{cutmix} & 0 & 97.10 & 97.62 \\
      FMix~\cite{harris2020fmix} & 6$^\dagger$ & 96.38 & - \\
      SaliencyMix~\cite{uddin2020saliencymix} & 6$^\dagger$ & 97.24 & - \\
      \midrule
      \name & 0 & 97.60 & 97.93 \\
      \name+ & 6 & \textbf{98.10} & \textbf{98.47} \\
      \bottomrule
      \end{tabular}
   \end{threeparttable}
   \vspace{-10pt}
\end{table}

The top-1 test accuracy comparisons are shown in Tab.~\ref{tab: cifar10_results}. We compare the results of our method with CutMix~\cite{cutmix}, and some saliency-guided mixing augmentations~\cite{kim2020puzzlemix,harris2020fmix,uddin2020saliencymix}, as well as some automated augmentation methods~\cite{autoaugmentcubuk,cubuk2020randaugment,fasteraahataya2019, fastaalim}. Our proposed \name\ augmentation outperforms CutMix~\cite{cutmix} by 0.5\% and it even outperforms the automated augmentation method AutoAugment~\cite{autoaugmentcubuk} by 0.28\% with WideResNet-28-10. It is worth noting that \name\ does not introduce any additional computation cost, while most saliency-guided or automated augmentation methods take additional cost to promote the performance.

\subsubsection{Experiments on CIFAR-100}
\label{section: cifar100}
The CIFAR-100 dataset has the same number of images as CIFAR-10 but it contains 100 classes. We apply our method \name\ on the WideResNet-28-10 and Shake-Shake (26 2x96d) network. We use the same settings and hyper-parameters as the CIFAR-10 dataset to train WideResNet-28-10 and Shake-Shake (26 2x96d). Tab.~\ref{tab: cifar100_results} shows the CIFAR-100 performance comparisons of our proposed \name\ with other cutting method~\cite{cutout2017}, mixing method~\cite{cutmix,kim2020puzzlemix,uddin2020saliencymix} and automated augmentations. We observe that \name\ outperforms CutMix~\cite{cutmix} by 0.87\%. Compared to the automated augmentations, it surpasses AutoAugment~\cite{autoaugmentcubuk} by 1.40\% and RandAugment~\cite{cubuk2020randaugment} by 1.01\%. 

\begin{table}[t!]
   \centering
   \caption{Top-1 test accuracy rate (\%) on CIFAR-100 classification with WideResNet-28-10 and Shake-Shake (26 2x96d).} 
   \label{tab: cifar100_results}
   \begin{threeparttable}\small
         \begin{tabular}{lccc}
         \toprule
         \textbf{Method} & \textbf{Cost (GHs)} & \textbf{WRS28-10} & \textbf{SS-2$\times$96d} \\ 
         \midrule
         Baseline & - & 81.20 & 82.95\\
         AA~\cite{autoaugmentcubuk} & 5000 & 82.91 & \textbf{85.72} \\
         Fast AA~\cite{fastaalim} & 3.5 & 82.70 & 85.40 \\
         PBA~\cite{pba} & 5 & 83.27 & 84.69 \\
         RA~\cite{cubuk2020randaugment} & 0 & 83.30 & - \\
         % AdvAA & 84.51 & 85.90 \\
         Faster AA~\cite{fasteraahataya2019} & 0.23 & 82.20 & 84.40 \\
         DADA~\cite{li2020dada} & 0.2 & 82.50 & 84.70 \\
         \midrule
         Cutout~\cite{cutout2017} & 0 & 81.59 & 84.0 \\
         CutMix~\cite{cutmix} & 0 & 83.40 & 85.0 \\
         FMix~\cite{harris2020fmix} & 6$^\dagger$ & 82.03 & - \\
         SaliencyMix~\cite{uddin2020saliencymix} & 6$^\dagger$ & 83.44 & - \\
         Puzzle Mix~\cite{kim2020puzzlemix} &12$^\dagger$ & 84.05 & - \\
         \midrule
         \name & 0 & 84.31 & 85.26 \\
         \name+ & 6 & \textbf{85.23} & 85.60 \\
         \bottomrule
         \end{tabular}
      \end{threeparttable}
   \end{table}
   
\begin{table}[t!]
   \centering
   \caption{Top-1 test accuracy rate (\%) on ImageNet classification with ResNet-50 and ResNet-101 networks.}
   \label{tab: imagenet_results}
   \begin{threeparttable}\small
         \begin{tabular}{lccc}
         \toprule
         \textbf{Method} & \textbf{Cost (GHs)} & \textbf{ResNet-50} & \textbf{ResNet-101}  \\ 
         \midrule
         Baseline & - & 76.31 & 78.13 \\
         AA~\cite{autoaugmentcubuk} & 15,000 & 77.63 & - \\
         FastAA~\cite{fastaalim} & 450 & 77.60 & - \\
         OHL-AA~\cite{lin2019online} & 625$^\dagger$ & 78.93 & - \\
         RA~\cite{cubuk2020randaugment} & 0 & 77.60 & - \\
         % AdvAA & 79.40 & - \\
         Faster AA~\cite{fasteraahataya2019} & 2.3 & 76.50 & - \\
         DADA~\cite{li2020dada} & 1.3 & 77.50 & - \\
         \midrule
         Cutmix~\cite{cutmix} & 0 & 78.60 & 79.83 \\
         SaliencyMix~\cite{uddin2020saliencymix} & 280$^\dagger$ & 78.74 & 79.91 \\
         Puzzle Mix~\cite{kim2020puzzlemix} & 576$^\dagger$ & 77.51 & - \\
         \midrule
         \name & 0 & \textbf{79.00} & \textbf{80.54} \\
         % \name+ &   & \\
         \bottomrule
         \end{tabular}
      \end{threeparttable}
   \end{table}  

\begin{table*}[thbp]
   \centering
   \caption{Generalization ability comparisons on object detection between \name\ and CutMix~\cite{cutmix}. The experiments are performed on two frameworks of SSD~\cite{liu2016ssd} and Faster-RCNN~\cite{ren2015faster} on both MS-COCO~\cite{COCO} and Pascal VOC~\cite{everingham2010pascal} datasets.}
   \label{tab: coco_results}
   \small
         \begin{tabular}{l|c|c|c|c|c}
         \hline
         \multirow{3}*{\textbf{Backbone}} & \multirow{3}*{\textbf{\begin{tabular}{c}
         ImageNet-Cls\\
            Top-1 ACC(\%)
         \end{tabular}}} & \multicolumn{2}{c|}{\textbf{MS-COCO Detection}} & \multicolumn{2}{c}{\textbf{Pascal VOC Detection}} \\
         \cline{3-6}
         && SSD & Faster-RCNN & {SSD} & Faster-RCNN \\ 
         && mAP(\%) & mAP(\%) & mAP(\%) & mAP(\%) \\
         \hline
         ResNet-50 & 76.1 & 25.1 & 38.1 & 75.6 & 81.0 \\
         Cutmix~\cite{cutmix} & 78.6 & 24.9 & 38.2 & 76.1 & 81.9 \\
         \hline
         \textbf{\name} & \textbf{79.0} & \textbf{25.5} &  \textbf{38.4} & \textbf{77.3} & \textbf{82.0} \\
         \hline
         \end{tabular}
         \vspace{-5pt}
   \end{table*}

\subsubsection{Experiments on ImageNet}
ImageNet~\cite{russa2015imagenet} is a challenging and widely used dataset for image classification. It contains 1.2 million training images and 50,000 validation images with 1,000 classes. The input image size is set as $224 \times 224$. We train our method with the networks of ResNet-50 and ResNet-101~\cite{deep} for 300 epochs. We set the batch size as 512, the initial learning rate as 0.5, and the weight decay as $4 \times 10^{-5}$. The learning rate decays with the cosine annealing schedule.
The ImageNet results are shown in Tab.~\ref{tab: imagenet_results}. With the ResNet-50 network, the performance of \name\ surpasses CutMix~\cite{cutmix} by 0.4\% and Puzzle Mix~\cite{kim2020puzzlemix} by 1.49\%. It outperforms the automated ones, AutoAugment~\cite{autoaugmentcubuk} by 1.37\%,  Faster AA~\cite{fasteraahataya2019} by 2.5\%. It is worth noting that AutoAugment needs the additional computation cost of 15,000 GPU hours while \name\ does not introduce any additional cost. For ResNet-101, the performance of \name\ exceeds the performance of CutMix by 0.71\%, which achieves the top-1 accuracy rate of 80.54\%.

\subsection{Evaluation on Object Detection}
\label{section: detection}
For evaluating the generalization ability of our method, we use the \name\ pre-trained ResNet-50~\cite{deep} model as the backbone network of two object detection frameworks, \ie Faster RCNN~\cite{ren2015faster} and SSD~\cite{liu2016ssd}. We perform the experiments on both MS-COCO~\cite{COCO} and Pascal VOC~\cite{everingham2010pascal} datasets. All the experiments are based on the object detection toolkit MMDetection~\cite{chen2019mmdetection}. For SSD training, the input image is resized to $300 \times 300$. The batch size is set as 64 for two datasets. It takes 24 epochs in total. Both VOC2007 and VOC2012 \texttt{trainval} (VOC07+12) are used for training, and the models are evaluated on the VOC 2007 benchmark. For Faser-RCNN training, the image scale is set as $(1333, 800)$ for MS-COCO and $(1000, 600)$ for Pascal VOC. It takes 12 epochs in total for MS-COCO and 4 epochs for Pascal VOC respectively. For all the other training hyper-parameters, we just follow the default settings defined in MMDetection.

As shown in Tab.~\ref{tab: coco_results}, our \name\ shows great generalization ability under several object detection evaluation settings. Especially on the lightweight framework SSD, \name\ shows notable mAP promotion over the baseline network, 0.4\% mAP on MS-COCO and 1.7\% mAP on Pascal VOC.

% \begin{table*}[thbp]
% \centering
% \caption{Our results on COCO detection} 
% \label{tab: coco_results}
% \small
%     \begin{tabular}{l|c|c|c|c|c|c|c}
%     \hline
%     \multirow{3}*{\textbf{Backbone}} & \multirow{3}*{\textbf{\begin{tabular}{c}
%     ImageNet-Cls\\
%         Top-1 ACC (\%)
%     \end{tabular}}} & \multicolumn{4}{c|}{\textbf{MS-COCO Detection}} & \multicolumn{2}{c}{\textbf{Pascal VOC Detection}} \\
%     \cline{3-8}
%     && \multicolumn{2}{c|}{SSD} & \multicolumn{2}{c|}{{Faster-RCNN}} & {SSD} & Faster-RCNN \\ 
%     \cline{3-8}
%     && mAP & mAP$^m$ & mAP & mAP$^s$ & mAP & mAP \\
%     \hline
%     ResNet-50 & 76.1 & 25.1 & 26.9 & 38.1 & 22.2 & 75.6 & 81.0 \\
%     Cutmix~\cite{cutmix} & 78.6 & 24.9 & 26.5 & 38.2 & 22.7 & 76.1 & 81.9 \\
%     \hline
%     \textbf{\name} & \textbf{79.0} & \textbf{25.5} & \textbf{27.9} & \textbf{38.4} & \textbf{23.2} & \textbf{77.3} & \textbf{82.0} \\
%     \hline
%     \end{tabular}
% \end{table*}

\subsection{Ablation Study and Analysis}
\label{section: ablation_study}
In this section, we perform a series of ablation studies and analysis about \name\ and other mixing-based augmentations.
We first study the advantage of resizing over cutting on preserving the source image information in Sec.~\ref{section: resize and cut}. Then we combine RandAugment~\cite{cubuk2020randaugment} with \name\ and further promote the performance in Sec.~\ref{section: randaug_study}. Next, we explore several settings of resizing scale rates in Sec.~\ref{section: resize scale study}. Finally, we analyze the differences between \name\ and other mixing-based augmentations in Sec.~\ref{section: analysis of mixing}.

\begin{table}[t!]
\centering
\caption{Comparisons of the effects between resizing and cropping on the half input resolution training. The shown results are all the top-1 accuracies (\%) on the validation set. The ``Train'' and ``Val'' column indicate the strategies of obtaining half-resolution input images for training and validation respectively. ``RandCrop'' means randomly cropping a patch from the image and ``Resize'' means resizing the whole image to a smaller patch. ``CenterCrop'' means cropping a patch at the center of the testing image.}
\label{tab: differ_crop-resize} 
\begin{threeparttable}
\resizebox{\linewidth}{!}{
      \begin{tabular}{cccccc}
      \toprule
      \multirow{2}*{\textbf{Row}} &      \multirow{2}*{\textbf{Train}} & \multirow{2}*{\textbf{Val}} & \textbf{CIFAR-10} & \textbf{CIFAR-100} & \textbf{ImageNet} \\
      & & & \textbf{WRS28-10} & \textbf{WRS28-10} & \textbf{ResNet-50} \\
      \midrule
      Baseline & - & - & 96.13 & 81.20 & 76.31 \\
      \midrule
      (1) & RandCrop & Resize & 71.80 & 35.84 & 63.59 \\
      (2) & RandCrop & CenterCrop & 90.10 & 66.70 & 58.58  \\
      (3) & Resize & Resize & \textbf{92.06} & \textbf{71.90} & \textbf{63.85} \\
      \bottomrule
      \end{tabular}}
   \end{threeparttable}\vspace{-15pt}
\end{table} 

\vspace{-10pt}
\subsubsection{Cutting \emph{vs}\onedot Resizing on Information Preserving}
\label{section: resize and cut}
We get the conclusion from Sec.~\ref{section: how_obtain} that cutting a patch from the source image may cause the problem of object information missing. To further verify the different effects of cutting and resizing on data mixing, we implement the comparison experiments under half input resolution settings. Specifically, during training, the input image is processed to a half-resolution one by randomly cropping a patch from the image or resizing the image to a half size. The images for validation are processed to the half sizes as well. The half-resolution experiments aim at comparing the information preserving abilities between cutting and resizing.

As shown in Tab.~\ref{tab: differ_crop-resize}, processing the training images into the half-resolution ones by resizing shows evident advantages over cutting. When the training images are processed by cutting, no matter the testing images are processed by resizing or cutting at the image center, the final performance cannot surpass that with resizing the training images. This further demonstrates that for obtaining a patch from the image, the manner of resizing preserves more effective information than cutting.
% the randcrop and centercrop mean randomly cropping the source image to 0.5 scale patch when training and making centercrop on the test set. The randcrop and resize mean randomly cropping the source image in 0.5 ratio and resizing the test set to 0.5 ratio. The resize and resize represent resizing all the image to 0.5 ratio when training and testing. We observe that the randcrop image manner obtains a worse performance than the resized image, which proves our hypothesis that cropping image may lead to label misallocation and object information missing.

\subsubsection{Effect of RandAugment on \name}
\label{section: randaug_study}
We are the first to study the effect of automated data augmentation on mixing data augmentation by combining \name\ with RandAugment. To verify the impact of the position relationship between \name\ and RandAugment on the training performance, we place the RandAugment operations before and after \name respectively. We perform the experiment on the CIFAR-100 dataset with WideResNet-28-10, and all the hyperparameter settings are the same as that in Sec.~\ref{section: cifar100}. As shown in Tab.~\ref{tab: randaug_effect}, the performance of putting RandAugment before \name\ is worse than using \name\ individually. This indicates that performing RandAugment on two images independently before mixing leads the two images to different patterns, which destroys the naturality of the mixed image and hinders the network learning. While RandAugment is performed after the images are mixed, the performance of \name\ obtains further improvement. It can be concluded that adding RandAugment after \name\ is a stronger augmentation pipeline to obtain better performance. It is worth noting that both \name\ and RandAugment do not introduce any additional computation cost.

\begin{table}[t!]
   \centering
   \caption{The results of different RandAugment placing positions on CIFAR-100 with WideResNet-28-10. ``Before'' means placing RandAugment operations before \name, and ``After'' means placing RandAugment after \name.}
   \label{tab: randaug_effect}
   \begin{threeparttable}\small
         \begin{tabular}{lcccc}
         \toprule
         \textbf{Position} & \textbf{Baseline} & \textbf{\name} & \textbf{Before} & \textbf{After} \\
         \midrule
         \textbf{Top-1(\%)} & 81.2 & 84.31 & 83.47 & \textbf{84.59} \\
         \bottomrule
         \end{tabular}
      \end{threeparttable}
      \vspace{-10pt}
   \end{table}

When equipped with RandAugment~\cite{cubuk2020randaugment} and the batch augmentation strategy~\cite{hoffer2020augment, zhang2019adversarialaa, lin2019online} (the enlarging scale is set as 2 in our experiments), \name+ achieves top-1 accuracy rates of 98.10\% with WideResNet-28-10 and 98.47\% with Shake-Shake (26 2x96d) on CIFAR-10 in Tab.~\ref{tab: cifar10_results}. And \name+ also achieves the new state-of-the-art performance of \textbf{85.23\%} on CIFAR-100 with WideResNet-28-10 in Tab.~\ref{tab: cifar100_results}.

\begin{table}[thbp]
\centering
\caption{Comparison with different resizing scale ranges on CIFAR-100 with WideResnet-28-10.}
\label{tab: effect_scale}
\resizebox{\linewidth}{!}{
\begin{threeparttable}
      \begin{tabular}{lccccc}
      \toprule
      \textbf{Range} & \textbf{Baseline} & \textbf{0.1-0.9} & \textbf{0.1-0.8} & \textbf{0.1-0.7}  & \textbf{0.2-0.8}\\  
      \midrule
      \textbf{Top-1(\%)} & 81.20 & 83.91 & \textbf{84.31} & 83.72 & 83.70 \\
      \bottomrule
      \end{tabular}
   \end{threeparttable}}
   \vspace{-15pt}
\end{table}

\subsubsection{Studying Resizing Scales}
\label{section: resize scale study}
In this section, we study the settings of the resizing scale ratio. Since the scale ratio $\tau$ is randomly sampled from the uniform distribution $U(\alpha, \beta)$, we set different $\alpha$ and $\beta$ to limit the range of ratio $\tau$. Tab.~\ref{tab: effect_scale} shows the results of different $\alpha$ and $\beta$ settings. All the experiments are performed on CIFAR-100 with WideResNet-28-10, and all the settings are the same as that in Sec.~\ref{section: cifar100}. We observe that when setting $\alpha$ as 0.1 and $\beta$ as 0.8 obtains the best performance, which is adopted to all the experiments with \name.

\subsubsection{Analysis on Different Mixing Methods}
\label{section: analysis of mixing}
\begin{figure}[t]
   \begin{center}
      \includegraphics[width=0.8\linewidth]{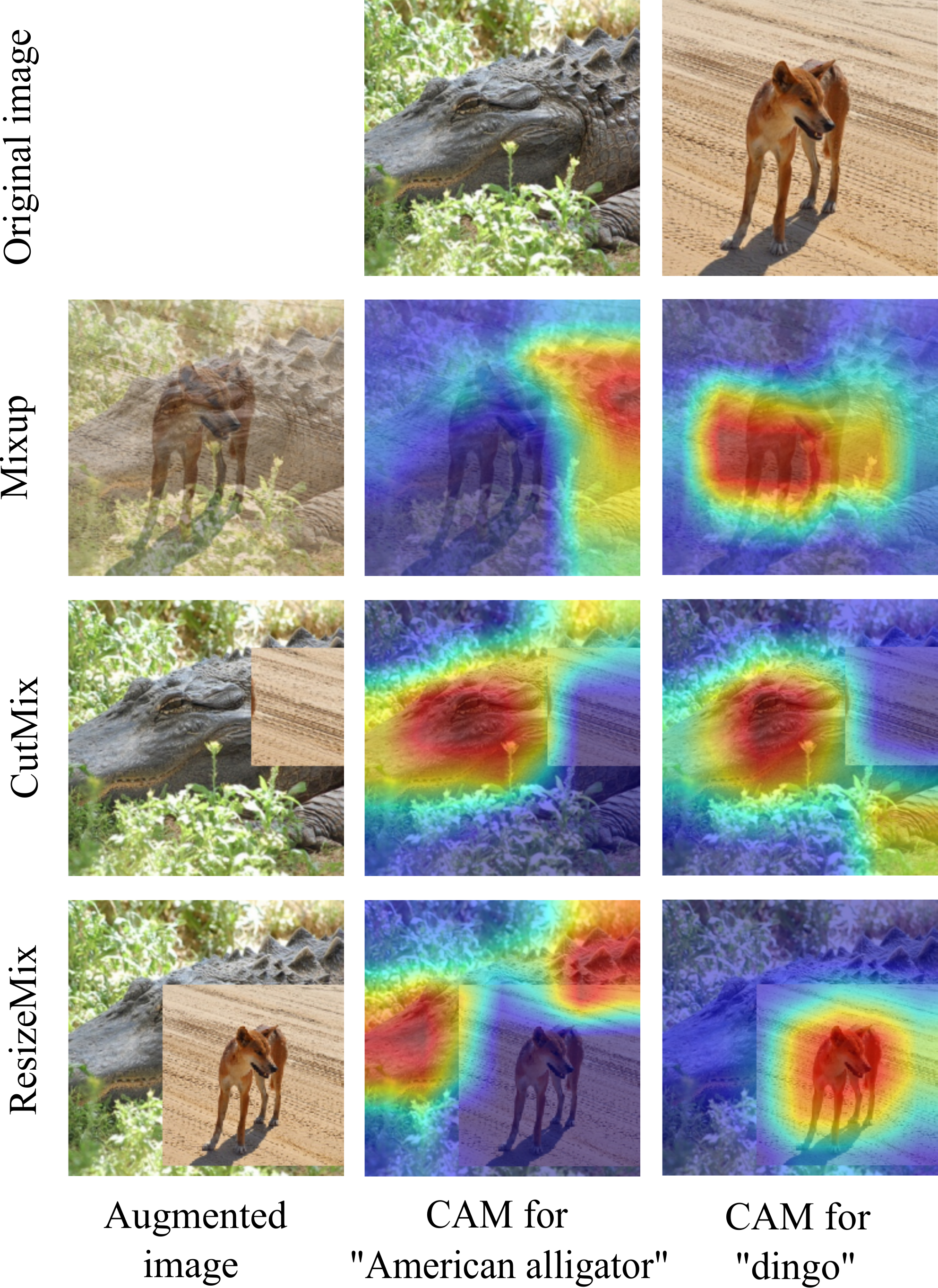}
   \end{center}
   \vspace{-15pt}
   \caption{CAM visualization on "American alligator" and "dingo" using different augmentations.\vspace{-15pt}
   }
   \label{fig: compare_mix}
\end{figure}

We visualize the CAM~\cite{zhou2016learningcam} heatmaps of images mixed with different methods. As shown in Fig.~\ref{fig: compare_mix}, the first row are the original images, and the first column on the left are the mixed images of various mixing methods including Mixup~\cite{mixup},  CutMix~\cite{cutmix}, and \name. And the next two columns show the CAM heatmaps of categories "American alligator" and "dingo" respectively. 

We observe that though the Mixup-generated image contains the informations of both categories, the mixed image in unnatural campared with real-life images. %The Cutout method occludes part of region of iamge, and the network only can learn the information of "American alligator". 
CutMix pastes a random patch of the source image into another image, but the patch is more likely to contain no information of "dingo", which leads to the problem of label misallocation. The network cannot locate the region corresponding to the label "dingo" and this will mislead the network learning. However, \name\ obtains the patch preserving all the information of the source image "dingo", which effectively eliminates label misallocation.

\section{Conclusion}
In this paper, we systematically study the CutMix-based data augmentation methods, and find that the saliency information of mixing data is not so necessary. Moreover, we conclude that the cutting-based data mixing strategies cannot avoid label misallocation and object information missing simultaneously. To tackle the two intractable problems, we propose an effective method, namely \name, which directly resizes the image to a small patch and mixes it with another image. The proposed method shows evident advantages over previous methods on various image classification and object detection benchmarks.

\section*{Acknowledgement}
We thank Liangchen Song for the discussion and assistance.

{\small
\bibliographystyle{ieee_fullname}
\bibliography{egbib}

\begin{thebibliography}{10}\itemsep=-1pt

\bibitem{DBLP:journals/corr/Alistarh0TV16}
Dan Alistarh, Jerry Li, Ryota Tomioka, and Milan Vojnovic.
\newblock {QSGD:} randomized quantization for communication-optimal stochastic
  gradient descent.
\newblock {\em arXiv:1610.02132}, 2016.

\bibitem{antoniou2017data}
Antreas Antoniou, Amos Storkey, and Harrison Edwards.
\newblock Data augmentation generative adversarial networks.
\newblock {\em arXiv:1711.04340}, 2017.

\bibitem{bergstra2012random}
James Bergstra and Yoshua Bengio.
\newblock Random search for hyper-parameter optimization.
\newblock {\em JMLR}, 2012.

\bibitem{cai2018proxylessnas}
Han Cai, Ligeng Zhu, and Song Han.
\newblock Proxyless{NAS}: Direct neural architecture search on target task and
  hardware.
\newblock In {\em ICLR}, 2019.

\bibitem{chen2019mmdetection}
Kai Chen, Jiaqi Wang, Jiangmiao Pang, Yuhang Cao, Yu Xiong, Xiaoxiao Li,
  Shuyang Sun, Wansen Feng, Ziwei Liu, Jiarui Xu, Zheng Zhang, Dazhi Cheng,
  Chenchen Zhu, Tianheng Cheng, Qijie Zhao, Buyu Li, Xin Lu, Rui Zhu, Yue Wu,
  and Dahua Lin.
\newblock Mmdetection: Open mmlab detection toolbox and benchmark.
\newblock {\em arXiv:1906.07155}, 2019.

\bibitem{DBLP:journals/pami/ChenPKMY18}
Liang-Chieh Chen, George Papandreou, Iasonas Kokkinos, Kevin Murphy, and Alan~L
  Yuille.
\newblock Deeplab: Semantic image segmentation with deep convolutional nets,
  atrous convolution, and fully connected crfs.
\newblock {\em TPAMI}, 2017.

\bibitem{choe2019attention}
Junsuk Choe and Hyunjung Shim.
\newblock Attention-based dropout layer for weakly supervised object
  localization.
\newblock In {\em CVPR}, 2019.

\bibitem{autoaugmentcubuk}
Ekin~D Cubuk, Barret Zoph, Dandelion Mane, Vijay Vasudevan, and Quoc~V Le.
\newblock Autoaugment: Learning augmentation policies from data.
\newblock {\em CVPR}, 2018.

\bibitem{cubuk2020randaugment}
Ekin~D Cubuk, Barret Zoph, Jonathon Shlens, and Quoc~V Le.
\newblock Randaugment: Practical automated data augmentation with a reduced
  search space.
\newblock In {\em CVPR Workshops}, 2020.

\bibitem{dabouei2020supermix}
Ali Dabouei, Sobhan Soleymani, Fariborz Taherkhani, and Nasser~M Nasrabadi.
\newblock Supermix: Supervising the mixing data augmentation.
\newblock {\em arXiv:2003.05034}, 2020.

\bibitem{cutout2017}
Terrance DeVries and Graham~W Taylor.
\newblock Improved regularization of convolutional neural networks with cutout.
\newblock {\em arXiv:1708.04552}, 2017.

\bibitem{domhan2015speeding}
Tobias Domhan, Jost~Tobias Springenberg, and Frank Hutter.
\newblock Speeding up automatic hyperparameter optimization of deep neural
  networks by extrapolation of learning curves.
\newblock In {\em IJCAI}, 2015.

\bibitem{dozat2016incorporating}
Timothy Dozat.
\newblock Incorporating nesterov momentum into adam.
\newblock 2016.

\bibitem{everingham2010pascal}
Mark Everingham, Luc Van~Gool, Christopher~KI Williams, John Winn, and Andrew
  Zisserman.
\newblock The pascal visual object classes (voc) challenge.
\newblock {\em IJCV}, 2010.

\bibitem{fang2020fast}
Jiemin Fang, Yuzhu Sun, Kangjian Peng, Qian Zhang, Yuan Li, Wenyu Liu, and
  Xinggang Wang.
\newblock Fast neural network adaptation via parameter remapping and
  architecture search.
\newblock In {\em ICLR}, 2020.

\bibitem{fang2020densely}
Jiemin Fang, Yuzhu Sun, Qian Zhang, Yuan Li, Wenyu Liu, and Xinggang Wang.
\newblock Densely connected search space for more flexible neural architecture
  search.
\newblock In {\em CVPR}, 2020.

\bibitem{fang2020fna++}
Jiemin Fang, Yuzhu Sun, Qian Zhang, Kangjian Peng, Yuan Li, Wenyu Liu, and
  Xinggang Wang.
\newblock Fna++: Fast network adaptation via parameter remapping and
  architecture search.
\newblock {\em TPAMI}, 2020.

\bibitem{faster2015towards}
RCNN Faster.
\newblock Towards real-time object detection with region proposal networks.
\newblock {\em NeurIPS}, 2015.

\bibitem{gastaldi2017shake}
Xavier Gastaldi.
\newblock Shake-shake regularization.
\newblock {\em arXiv:1705.07485}, 2017.

\bibitem{2018dropblock}
Golnaz Ghiasi, Tsung-Yi Lin, and Quoc~V Le.
\newblock Dropblock: A regularization method for convolutional networks.
\newblock In {\em NeurIPS}, 2018.

\bibitem{gurumurthy2017adversarial}
Swaminathan Gurumurthy, Ravi Kiran~Sarvadevabhatla, and R Venkatesh~Babu.
\newblock Deligan: Generative adversarial networks for diverse and limited
  data.
\newblock In {\em CVPR}, 2017.

\bibitem{harris2020fmix}
Ethan Harris, Antonia Marcu, Matthew Painter, Mahesan Niranjan, and Adam
  Pr{\"u}gel-Bennett~Jonathon Hare.
\newblock Fmix: Enhancing mixed sample data augmentation.
\newblock {\em arXiv:2002.12047}, 2020.

\bibitem{fasteraahataya2019}
Ryuichiro Hataya, Jan Zdenek, Kazuki Yoshizoe, and Hideki Nakayama.
\newblock Faster autoaugment: Learning augmentation strategies using
  backpropagation.
\newblock {\em arXiv:1911.06987}, 2019.

\bibitem{he2016resnet}
Kaiming He, Xiangyu Zhang, Shaoqing Ren, and Jian Sun.
\newblock Deep residual learning for image recognition.
\newblock In {\em CVPR}, 2016.

\bibitem{deep}
Kaiming He, Xiangyu Zhang, Shaoqing Ren, and Jian Sun.
\newblock Deep residual learning for image recognition.
\newblock In {\em CVPR}, 2016.

\bibitem{hendrycks2019augmix}
Dan Hendrycks, Norman Mu, Ekin~D Cubuk, Barret Zoph, Justin Gilmer, and Balaji
  Lakshminarayanan.
\newblock Augmix: A simple data processing method to improve robustness and
  uncertainty.
\newblock {\em ICLR}, 2020.

\bibitem{pba}
Daniel Ho, Eric Liang, Xi Chen, Ion Stoica, and Pieter Abbeel.
\newblock Population based augmentation: Efficient learning of augmentation
  policy schedules.
\newblock In {\em ICML}, 2019.

\bibitem{hoffer2020augment}
Elad Hoffer, Tal Ben-Nun, Itay Hubara, Niv Giladi, Torsten Hoefler, and Daniel
  Soudry.
\newblock Augment your batch: better training with larger batches.
\newblock {\em CVPR}, 2020.

\bibitem{kim2020puzzlemix}
Jang-Hyun Kim, Wonho Choo, and Hyun~Oh Song.
\newblock Puzzle mix: Exploiting saliency and local statistics for optimal
  mixup.
\newblock {\em ICML}, 2020.

\bibitem{DBLP:journals/corr/KingmaB14}
Diederik~P. Kingma and Jimmy Ba.
\newblock Adam: {A} method for stochastic optimization.
\newblock In {\em ICLR}, 2015.

\bibitem{krizhevsky2009learning}
Alex Krizhevsky, Geoffrey Hinton, et~al.
\newblock Learning multiple layers of features from tiny images.
\newblock 2009.

\bibitem{li2020dada}
Yonggang Li, Guosheng Hu, Yongtao Wang, Timothy Hospedales, Neil~M Robertson,
  and Yongxing Yang.
\newblock Dada: Differentiable automatic data augmentation.
\newblock {\em arXiv:2003.03780}, 2020.

\bibitem{fastaalim}
Sungbin Lim, Ildoo Kim, Taesup Kim, Chiheon Kim, and Sungwoong Kim.
\newblock Fast autoaugment.
\newblock In {\em NeurIPS}, 2019.

\bibitem{lin2019online}
Chen Lin, Minghao Guo, Chuming Li, Xin Yuan, Wei Wu, Junjie Yan, Dahua Lin, and
  Wanli Ouyang.
\newblock Online hyper-parameter learning for auto-augmentation strategy.
\newblock In {\em ICCV}, 2019.

\bibitem{COCO}
Tsung{-}Yi Lin, Michael Maire, Serge~J. Belongie, James Hays, Pietro Perona,
  Deva Ramanan, Piotr Doll{\'{a}}r, and C.~Lawrence Zitnick.
\newblock Microsoft {COCO:} common objects in context.
\newblock In {\em ECCV}, 2014.

\bibitem{lin2017focal}
Tsung-Yi Lin, Priya Goyal, Ross Girshick, Kaiming He, and Piotr Doll{\'a}r.
\newblock Focal loss for dense object detection.
\newblock In {\em ICCV}, 2017.

\bibitem{liu2018darts}
Hanxiao Liu, Karen Simonyan, and Yiming Yang.
\newblock {DARTS}: Differentiable architecture search.
\newblock In {\em ICLR}, 2019.

\bibitem{liu2016ssd}
Wei Liu, Dragomir Anguelov, Dumitru Erhan, Christian Szegedy, Scott Reed,
  Cheng-Yang Fu, and Alexander~C Berg.
\newblock Ssd: Single shot multibox detector.
\newblock In {\em ECCV}, 2016.

\bibitem{DBLP:conf/iclr/LoshchilovH17}
Ilya Loshchilov and Frank Hutter.
\newblock {SGDR:} stochastic gradient descent with warm restarts.
\newblock In {\em ICLR}, 2017.

\bibitem{peng2018jointlyadersarial}
Xi Peng, Zhiqiang Tang, Fei Yang, Rogerio~S Feris, and Dimitris Metaxas.
\newblock Jointly optimize data augmentation and network training: Adversarial
  data augmentation in human pose estimation.
\newblock In {\em CVPR}, 2018.

\bibitem{ren2015faster}
Shaoqing Ren, Kaiming He, Ross Girshick, and Jian Sun.
\newblock Faster r-cnn: Towards real-time object detection with region proposal
  networks.
\newblock In {\em NeurIPS}, 2015.

\bibitem{ronneberger2015u}
Olaf Ronneberger, Philipp Fischer, and Thomas Brox.
\newblock U-net: Convolutional networks for biomedical image segmentation.
\newblock In {\em LNCS}, 2015.

\bibitem{russa2015imagenet}
Olga Russakovsky, Jia Deng, Hao Su, Jonathan Krause, Sanjeev Satheesh, Sean Ma,
  Zhiheng Huang, Andrej Karpathy, Aditya Khosla, Michael Bernstein, et~al.
\newblock Imagenet large scale visual recognition challenge.
\newblock {\em IJCV}, 2015.

\bibitem{sandler2018mobilenetv2}
Mark Sandler, Andrew Howard, Menglong Zhu, Andrey Zhmoginov, and Liang-Chieh
  Chen.
\newblock Mobilenetv2: Inverted residuals and linear bottlenecks.
\newblock In {\em CVPR}, 2018.

\bibitem{selvaraju2017gradcam}
Ramprasaath~R Selvaraju, Michael Cogswell, Abhishek Das, Ramakrishna Vedantam,
  Devi Parikh, and Dhruv Batra.
\newblock Grad-cam: Visual explanations from deep networks via gradient-based
  localization.
\newblock In {\em ICCV}, 2017.

\bibitem{2017hide}
Krishna~Kumar Singh and Yong~Jae Lee.
\newblock Hide-and-seek: Forcing a network to be meticulous for
  weakly-supervised object and action localization.
\newblock In {\em ICCV}, 2017.

\bibitem{srivasdropout2014}
Nitish Srivastava, Geoffrey Hinton, Alex Krizhevsky, Ilya Sutskever, and Ruslan
  Salakhutdinov.
\newblock Dropout: a simple way to prevent neural networks from overfitting.
\newblock {\em JMLR}, 2014.

\bibitem{summers2019improved}
Cecilia Summers and Michael~J Dinneen.
\newblock Improved mixed-example data augmentation.
\newblock In {\em WACV}, 2019.

\bibitem{takahashi2018ricap}
Ryo Takahashi, Takashi Matsubara, and Kuniaki Uehara.
\newblock Ricap: Random image cropping and patching data augmentation for deep
  cnns.
\newblock In {\em ACML}, 2018.

\bibitem{uddin2020saliencymix}
AFM Uddin, Mst Monira, Wheemyung Shin, TaeChoong Chung, Sung-Ho Bae, et~al.
\newblock Saliencymix: A saliency guided data augmentation strategy for better
  regularization.
\newblock {\em arXiv:2006.01791}, 2020.

\bibitem{cutmix}
Sangdoo Yun, Dongyoon Han, Seong~Joon Oh, Sanghyuk Chun, Junsuk Choe, and
  Youngjoon Yoo.
\newblock Cutmix: Regularization strategy to train strong classifiers with
  localizable features.
\newblock In {\em ICCV}, 2019.

\bibitem{zagoruyko2016wide}
Sergey Zagoruyko and Nikos Komodakis.
\newblock Wide residual networks.
\newblock {\em arXiv:1605.07146}, 2016.

\bibitem{mixup}
H Zhang, M Cisse, Y Dauphin, and D Lopez-Paz.
\newblock mixup: Beyond empirical risk minimization.
\newblock {\em ICLR}, 2018.

\bibitem{zhang2019adversarialaa}
Xinyu Zhang, Qiang Wang, Jian Zhang, and Zhao Zhong.
\newblock Adversarial autoaugment.
\newblock {\em ICLR}, 2020.

\bibitem{erazing2020}
Zhun Zhong, Liang Zheng, Guoliang Kang, Shaozi Li, and Yi Yang.
\newblock Random erasing data augmentation.
\newblock In {\em AAAI}, 2020.

\bibitem{zhou2016learningcam}
Bolei Zhou, Aditya Khosla, Agata Lapedriza, Aude Oliva, and Antonio Torralba.
\newblock Learning deep features for discriminative localization.
\newblock In {\em CVPR}, 2016.

\bibitem{zoph2018learning}
Barret Zoph, Vijay Vasudevan, Jonathon Shlens, and Quoc~V Le.
\newblock Learning transferable architectures for scalable image recognition.
\newblock In {\em CVPR}, 2018.

\end{thebibliography}
}

\appendix
\section{Appendix}
\subsection{Details of Obtaining Salient and Non-salient Regions}
We first obtain the heatmap of the input image by using the Grad-CAM~\cite{selvaraju2017gradcam} module. We denote the activation values of the heatmap as $A$. Two thresholds $t_u$ and $t_l$ are set as the maximum and minimum activation values of $A$, 
\begin{equation}
   t_u = max(A)\text{,}\quad
   t_l = min(A)\text{.}
\end{equation}
As there are many pixels which hold the activation values of the maximum or minimum values, we get the sets of salient and non-salient pixel coordinates $C_s$ and $C_{ns}$ as
\begin{equation}
   \begin{aligned}
   C_s &= \{(x, y) | A(x, y) \ge t_{u}\}\text{,}\\
   C_{ns} &= \{(x, y) | A(x, y) \le t_{l}\}\text{,}
   \end{aligned}
\end{equation}
where $A(x, y)$ denotes the activation value of the heatmap at the coordinate of $(x, y)$.

% For the image, we obtain the salient and non-salient region when got the width $W_P$ and the height $H_P$ of the the source patch. 
We obtain the salient region $W_P \times H_P$ of a image as follows. We first randomly sample a coordinate $(x_c, y_c)$ as the geometry center of the region from $C_s$, \ie, $(x_c, y_c) \in C_s$. Then we calculate the boundaries of $R_s$ as
\begin{equation}
  \begin{aligned}
  x_l &= \lceil x_c - \frac{W_P}{2} \rceil \text{,}\quad &x_r &= \lfloor x_c + \frac{W_P}{2} \rfloor \text{,} \\ 
  y_b &= \lceil y_c - \frac{H_P}{2} \rceil \text{,}\quad &y_t &= \lfloor y_c + \frac{H_P}{2} \rfloor \text{,}  
  \end{aligned}
\end{equation}
where $x_l$, $x_r$, $y_b$, $y_t$ denote the left, right, bottom and top boundary of the salient region $R_s$. 
\begin{figure*}[t]
   \begin{center}
      \includegraphics[width=\linewidth]{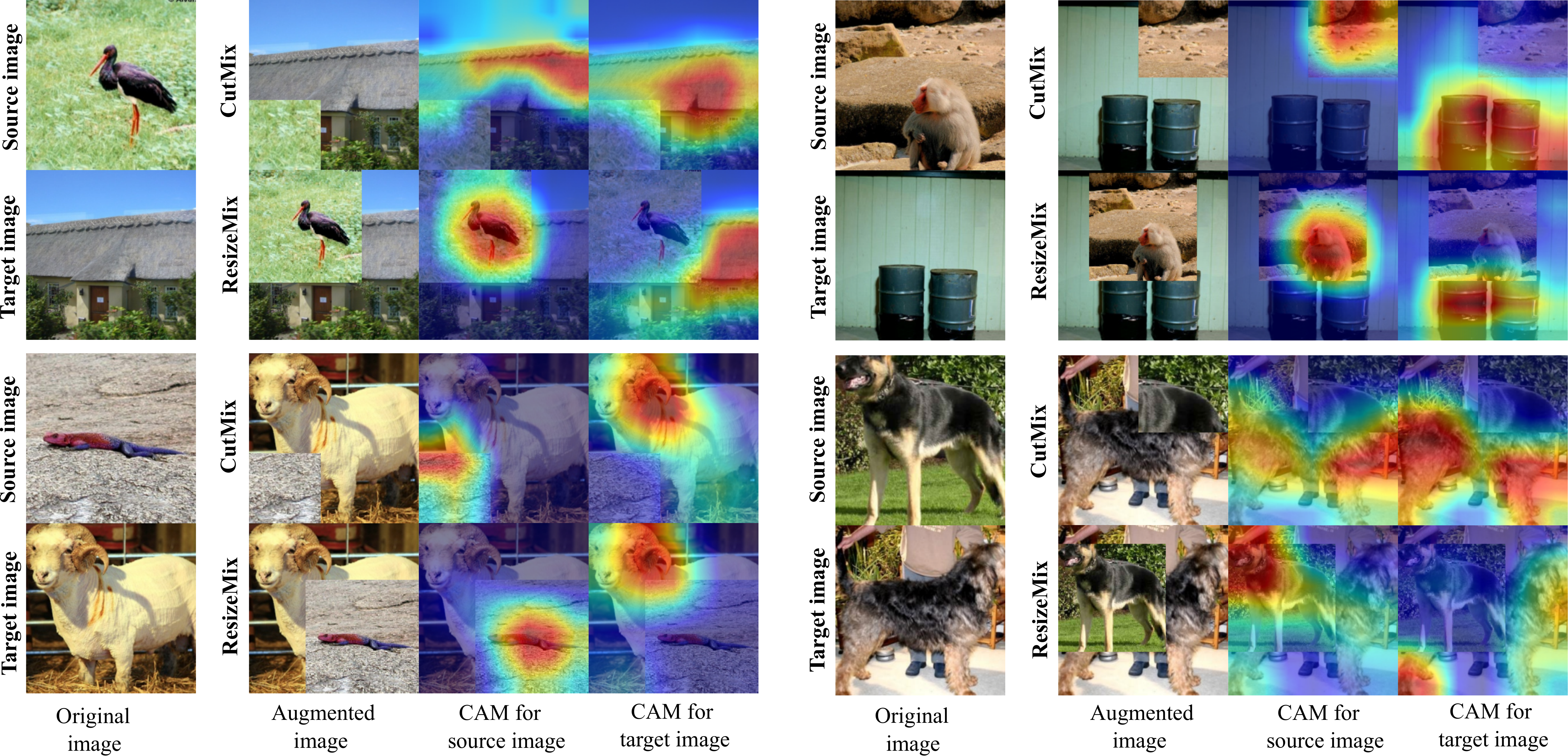}
   \end{center}
   \vspace{-10pt}
   \caption{More visualization comparisons between CutMix and ResizeMix.}
   \label{fig: compare_results}
\end{figure*}
Finally, we adjust these boundaries to guarantee the whole region is within the image.
For $x_l$ and $x_r$,
\begin{equation}
  \begin{aligned}
    %  &x_r = x_l + W_P\text{,} \\
     if \  x_l \leq 0\text{,}
     &\begin{cases}
        x_l &= 0\text{,}  \\
        x_r &= W_P\text{,} 
     \end{cases} \\
     if \  x_r \ge W\text{,}
     &\begin{cases}
        x_l &= W-W_P\text{,}  \\
        x_r &= W\text{;} 
     \end{cases}
  \end{aligned}
\end{equation}
For $y_b$ and $y_t$,
\begin{equation}
  \begin{aligned}
     if \  y_b \leq 0\text{,}
     &\begin{cases}
        y_b &= 0\text{,}  \\
        y_t &= H_P\text{,} 
     \end{cases} \\
     if \  y_t \ge H\text{,}
     &\begin{cases}
        y_b &= H-H_P\text{,}  \\
        y_t &= H\text{,} 
     \end{cases}
  \end{aligned}
\end{equation}
where $W$ and $H$ denote the width and height of the target image. The non-salient region $R_{ns}$ can be obtained in the same way.

\end{document}